\documentclass[final,5p,times,twocolumn,authoryear]{elsarticle}

\usepackage{amssymb}
\usepackage{algorithm}
\usepackage{algpseudocode}
\usepackage{graphicx}
\usepackage{url}
\newcommand{\ie}{{\emph{i.e.}} }

\usepackage{subfigure}
\usepackage{amsmath}

\usepackage{amssymb}
\usepackage{xcolor}
\usepackage{tabularx} 
\usepackage{multirow}
\usepackage{pbox}
\usepackage{booktabs}

\usepackage{amssymb}
\usepackage{amsfonts}
\usepackage{bbm}
\usepackage{graphicx}
{\begin{list}               
    {$\bullet$ \hfill}{
        \setlength{\leftmargin}{\parindent}
        \setlength{\parsep}{0.04\baselineskip}
        \setlength{\itemsep}{0.5\parsep}
        \setlength{\labelwidth}{\leftmargin}
        \setlength{\labelsep}{0em}}
    }
{\end{list}}

\providecommand{\eref}[1]{Eq. \eqref{#1}}  
\providecommand{\cref}[1]{Chapter~\ref{#1}}
\providecommand{\sref}[1]{Section~\ref{#1}}
\providecommand{\fref}[1]{Figure~\ref{#1}}
\providecommand{\tref}[1]{Table~\ref{#1}}

\providecommand{\norm}[1]{\lVert#1\rVert}

\renewcommand{\vec}[1]{\ensuremath{\boldsymbol{#1}}}
\providecommand{\mat}[1]{\ensuremath{\boldsymbol{#1}}}

\providecommand{\calC}{\mathcal{C}}
\providecommand{\calD}{\mathcal{D}}

\providecommand{\calL}{\mathcal{L}}

\providecommand{\calS}{\mathcal{S}}

\providecommand{\calX}{\mathcal{X}}

\providecommand{\calZ}{\mathcal{Z}}

\providecommand{\mI}{\mat{I}}

\providecommand{\vu}{\vec{u}}

\providecommand{\vx}{\vec{x}}
\providecommand{\vy}{\vec{y}}
\providecommand{\vz}{\vec{z}}

\providecommand{\vmu}{\vec{\mu}}

\journal{Neurocomputing}

\begin{document}

\begin{frontmatter}

\title{Unsupervised Contrastive Learning Using Out-Of-Distribution Data \\for Long-Tailed Dataset}

\author[1]{Cuong Manh Hoang}
\ead{cuonghoang@seoultech.ac.kr}

\author[2]{Yeejin Lee}
\ead{yeejinlee@seoultech.ac.kr}

\author[3]{Byeongkeun Kang\corref{cor1}}
\ead{byeongkeunkang@cau.ac.kr}
\cortext[cor1]{Corresponding author.}
\affiliation[1]{organization={Department of Electronic Engineering, Seoul National University of Science and Technology},
            addressline={232 Gongneung-ro, Nowon-gu}, 
            city={Seoul},
            postcode={01811}, 
            country={South Korea}}
\affiliation[2]{organization={Department of Electrical and Information Engineering, Seoul National University of Science and Technology},
            addressline={232 Gongneung-ro, Nowon-gu}, 
            city={Seoul},
            postcode={01811}, 
            country={South Korea}}
\affiliation[3]{organization={School of Electrical and Electronics Engineering, Chung-Ang University},
            addressline={84 Heukseok-ro, Dongjak-gu}, 
            city={Seoul},
            postcode={06974}, 
            country={South Korea}}

\begin{abstract}
This work addresses the task of self-supervised learning (SSL) on a long-tailed dataset that aims to learn balanced and well-separated representations for downstream tasks such as image classification. This task is crucial because the real world contains numerous object categories, and their distributions are inherently imbalanced. Towards robust SSL on a class-imbalanced dataset, we investigate leveraging a network trained using unlabeled out-of-distribution (OOD) data that are prevalently available online. We first train a network using both in-domain (ID) and sampled OOD data by back-propagating the proposed pseudo semantic discrimination loss alongside a domain discrimination loss. The OOD data sampling and loss functions are designed to learn a balanced and well-separated embedding space. Subsequently, we further optimize the network on ID data by unsupervised contrastive learning while using the previously trained network as a guiding network. The guiding network is utilized to select positive/negative samples and to control the strengths of attractive/repulsive forces in contrastive learning. We also distil and transfer its embedding space to the training network to maintain balancedness and separability. Through experiments on four publicly available long-tailed datasets, we demonstrate that the proposed method outperforms previous state-of-the-art methods.
\end{abstract}


\end{frontmatter}



\section{Introduction}
Self-supervised learning (SSL) is an important research topic because it enables learning representations without human-annotated labels. These learned representations are subsequently utilized in downstream tasks to reduce annotation efforts and improve accuracy~\citep{Neurocom2024Biswas, Neurocom2024Chai, Neurocom2025Yin, Neurocom2025Wen, EAAI2024LogacjovSelf, EAAI2023AkrimSelf}. For instance, once a backbone network is trained using SSL, a head network with fewer parameters can be appended to the frozen backbone and learned using a smaller dataset to address a downstream task. More specifically, to achieve robust accuracy given a small dataset, we can employ SSL to initially train a backbone network on unlabeled images from the web. Then, we append a head network to the backbone and train the head on the small target dataset. Accordingly, SSL is effective in various applications where data for target tasks are insufficient to learn effective representations. These applications include not only areas where data is hard to collect such as natural disasters or medical domains but also general tasks involving less-frequent objects such as tigers, irons, and operating scissors. Due to its significance, numerous researchers have investigated various methods to embed meaningful and balanced representations by training networks without human annotations~\citep{liu2023survey,tian2021divide}.

While previous SSL methods have demonstrated their effectiveness in learning representations without labels on balanced and curated datasets, they have achieved limited performance on imbalanced datasets~\citep{liu2022selfsupervised}. In~\tref{tab:balance_imbalance}, we also experimentally demonstrate the challenge caused by long-tailed data. We train models using a popular self-supervised learning method, SimCLR~\citep{chen2020simple}, on a balanced subset and a long-tailed subset of the CIFAR-10 and CIFAR-100 datasets~\citep{cui2019class}. Although both subsets contain the same total number of images, the distributions of the number of images for classes are different (balanced vs. imbalanced). For the long-tailed subset, we use an imbalance ratio of 100. As shown in the table, the models trained on the balanced subsets achieve higher overall accuracies and better balancedness (lower standard deviation) for both datasets. This limitation is critical because the real world contains numerous objects, and their distributions are inherently imbalanced~\citep{Neurocom2024Xiang, Neurocom2024Shi, Neurocom2024Wang, Neurocom2023Pang, EAAI2024FanCumulative, EAAI2024ZhangMultiple, EAAI2024LuAn}. Therefore, this paper addresses the problem of learning balanced and meaningful representations on long-tailed datasets.

\begin{table*}[!h]
\centering
\caption{Comparison of models trained on a balanced subset $D_b$ and a long-tailed subset $D_l$ from the CIFAR-10 and CIFAR-100 datasets~\citep{cui2019class}. Additional details on the evaluation metrics and protocols are provided in~\sref{sec:experimental_setting}.}
\label{tab:balance_imbalance}
\begin{minipage}{1\linewidth}
\centering
\begin{tabular}{>{\centering}m{0.15\textwidth}|>{\centering}m{0.1\textwidth}|>{\centering}m{0.08\textwidth} >{\centering}m{0.08\textwidth}  *{2}{>{\centering}m{0.08\textwidth}} >{\centering\arraybackslash}m{0.08\textwidth}} 
\toprule
Dataset & Subset & Many $\uparrow$ & Med. $\uparrow$ & Few $\uparrow$ & STD $\downarrow$ & All $\uparrow$  \\
\midrule
\multirow{2}{*}{CIFAR-10}& $D_b$ & 85.43 & 85.96 & 82.33 & 1.87 & 84.62  \\
										& $D_l$ & 82.40 & 73.91 & 70.19 & 5.11 & 75.34   \\
\midrule
\multirow{2}{*}{CIFAR-100} & $D_b$ &48.97 &48.69 &47.56 & 0.63 & 48.41  \\
											& $D_l$ & 51.50 & 45.58 & 45.96 & 2.71 & 47.65  \\
\bottomrule 
\end{tabular}
\end{minipage}
\end{table*}

Due to the importance of SSL on long-tailed datasets, researchers have explored various approaches. These approaches include applying strong regularization to rare samples~\citep{liu2022selfsupervised}, leveraging network pruning to discover tail samples~\citep{jiang2021self}, utilizing the memorization effect to find rare samples~\citep{zhou2022contrastive}, and using unlabeled out-of-distribution (OOD) datasets to rebalance tail samples~\citep{bai2023effectiveness}.

In long-tailed learning, one conventional approach rebalances the data distribution by re-sampling data from tail classes~\citep{resampling2002}. However, over-sampling tail samples often leads to overfitting due to the limited diversity of samples in those classes. To address this issue, another method employs under-sampling for head classes while over-sampling for tail classes with synthetic data~\citep{chawla2002smote}. While synthetic data mitigate overfitting, they are error-prone due to noise introduced during data synthesis~\citep{Drummond2003, cui2019class}. Consequently, researchers have investigated leveraging unlabeled in-distribution data to rebalance the data distribution~\citep{yang2020rethinking}. Although this approach improves long-tailed learning, collecting sufficient in-distribution data is challenging. As a result, recent studies have explored using unlabeled out-of-distribution data and demonstrated its effectiveness in addressing these challenges~\citep{wei2022open, bai2023effectiveness}.

Therefore, in this paper, we also investigate SSL on an unlabeled long-tailed dataset by leveraging unlabeled OOD data, considering prevalent online data without labels. However, different from~\citep{bai2023effectiveness}, we propose a two-step framework based on the hypothesis that using OOD data throughout the entire training might degrade accuracy because generating accurate pseudo-labels for the combined data is more challenging than that for only ID data. In the proposed framework, we firstly pre-train a network using both in-domain (ID) data and sampled OOD data to learn a balanced and well-separated embedding space. This step comprises tail sample discovery, OOD data sampling, and training to learn the embedding space. Subsequently, we utilize the pre-trained network to further optimize the network for the ID dataset. It involves employing the pre-trained network to select positive/negative samples and to control the strengths of attractive/repulsive forces in contrastive learning. Additionally, we distil the balanced and well-separated embedding space from the pre-trained network and transfer it to the training network.

The contributions of this paper are summarized as follows: (1) We present, to the best of our knowledge, the first SSL method for a long-tailed dataset that leverages a pre-trained network on OOD and ID data; (2) To learn a balanced and well-separated embedding space, we first train a network using the proposed pseudo semantic discrimination loss alongside a domain discrimination loss on both ID and carefully sampled OOD data; (3) We then leverage the pre-trained network to further optimize the network for an ID dataset. We propose to employ the pre-trained network to select positive/negative samples and to control the strengths of attractive/repulsive forces in contrastive learning. Additionally, we distil the balanced space from the pre-trained network and transfer it to the training network; (4) We demonstrate the effectiveness of the proposed method by freezing the further optimized network, appending a linear classifier to the frozen network, and training/evaluating on a long-tailed dataset.

\section{Related Work}
\subsection{Self-Supervised Learning with Long-Tailed Datasets}
\cite{yang2020rethinking} explored using self-supervised pre-training to mitigate label biases in class-imbalanced datasets. They demonstrated that classifiers pre-trained with self-supervised learning (SSL) consistently outperform their supervised learning counterparts for such datasets. Their SSL methods utilized rotation prediction~\citep{gidaris2018unsupervised} and Momentum Contrast (MoCo)~\citep{he2020momentum}.

To address biases caused by class imbalance, \cite{liu2022selfsupervised} introduced a re-weighted regularization method. This method begins by estimating the density of examples using a kernel estimation approach and subsequently applies strong regularization to the estimated rare samples using the Sharpness-Aware Minimization (SAM) technique~\citep{foret2021sharpnessaware}.
As an alternative method for estimating long-tail examples, \cite{jiang2021self} proposed using network pruning instead of kernel density estimation. This approach is based on the observation that network pruning has more impact on difficult-to-learn samples than easier ones. In detail, they introduced Self-Damaging Contrastive Learning (SDCLR), which emphasizes tail samples in a contrastive loss by utilizing a trained model and its pruned version.
To enhance the discovery of tail samples, \cite{zhou2022contrastive} leveraged the memorization effect (\ie the phenomenon that difficult-to-learn patterns are typically memorized after easier ones)~\citep{arpit2017acloser}. Building upon this observation, they introduced the Boosted Contrastive Learning (BCL) method, which emphasizes tail samples by employing more potent augmentation techniques.

Recently, \cite{bai2023effectiveness} explored utilizing external unlabeled OOD data to rebalance classes with rare samples. This approach is motivated by the fact that acquiring additional ID data is expensive while valuable for balancing long-tailed datasets. Hence, they estimated tail samples and added samples from external data to ID dataset to rebalance the long-tail distribution during training.

\subsection{Imbalanced Learning with Auxiliary Data}
\cite{jiang2021improving} explored a method that samples additional data from an external source to enhance SSL on imbalanced datasets. They introduced a novel unlabeled data sampling framework, Model-Aware K-center (MAK). It is designed to sample additional diverse data for tail classes while preventing the inclusion of out-of-distribution outliers. While \cite{jiang2021improving} demonstrated its effectiveness in improving representation quality and balancedness, this method assumes the presence of ID data within the external sources.

Since acquiring additional ID data is expensive, researchers have investigated using external datasets containing only OOD data to enhance imbalanced learning. \cite{wei2022open} presented a method that assigns open-set noisy labels to OOD data to rebalance label distributions for long-tailed datasets. It then trains a network using the labels like supervised learning. As mentioned earlier, \cite{bai2023effectiveness} recently introduced a method that enhances SSL on long-tailed datasets using OOD data. Considering the ease of collecting unlabeled online data, researchers have also explored using auxiliary data in various tasks, such as anomaly detection~\citep{hendrycks2018deep} and out-of-distribution data augmentation~\citep{lee2021removing}.

We investigate utilizing additional OOD data for self-supervised representation learning on long-tailed datasets similar to~\citep{bai2023effectiveness}. \cite{bai2023effectiveness} trained a network using OOD and ID data and directly employed the network for downstream tasks. In contrast to the previous work, we pre-train a network using OOD and ID data and then further optimize the network using ID data and the pre-trained network. It is to learn an optimal embedding function for ID data while maintaining the balanced and well-separated embedding space from the frozen network. We experimentally verify the effectiveness of the proposed method by applying it to two existing works, SimCLR~\citep{chen2020simple} and SDCLR~\citep{jiang2021self}. Different from~\citep{jiang2021improving}, we do not assume the presence of ID data within an external dataset nor use any additional ID data, similar to~\citep{wei2022open,bai2023effectiveness}.

\subsection{Supervised Learning with Long-Tailed Datasets} 
Supervised long-tailed learning methods have been extensively studied including re-balancing, representation learning, and prompt tuning. For re-balancing, \cite{park2022majority} introduced a minority over-sampling method that diversifies tail samples by pasting a tail sample onto an image from head classes. Later, \cite{gao2023enhancing} proposed an adaptive image-mixing method to synthesize semantically more reasonable images for tail classes. Recently, \cite{li2024feature} proposed to augment samples from tail classes by replacing parts of their feature maps with those from head classes.

Instead of generating synthetic images for minority classes, \cite{li2022long} introduced the Gaussian clouded logit adjustment method which assigns larger margins to tail classes than to head classes to enhance the separability of tail classes.  Later, \cite{chen2023transfer} presented a knowledge-transfer-based calibration method that estimates and utilizes importance weights for samples from rare classes. Another method, presented in~\citep{jin2023long}, proposed a self-heterogeneous integration framework that adopts and improves mixture-of-experts strategies.

With the advent of pre-trained models on web-scale datasets, \cite{dong2023lpt} presented a prompt tuning method for long-tailed classification tasks. To enhance generalization performance on rare classes, \cite{li2024improving} proposed the Gaussian neighborhood minimization prompt tuning method for visual prompt tuning. \cite{shi2024long} demonstrated that excessive fine-tuning of foundation models deteriorates performance on tail classes and introduced a lightweight fine-tuning method.

\section{Proposed Method}
\begin{figure*}[!h] 
\begin{minipage}{1\linewidth}
\centerline{\includegraphics[scale=0.55]{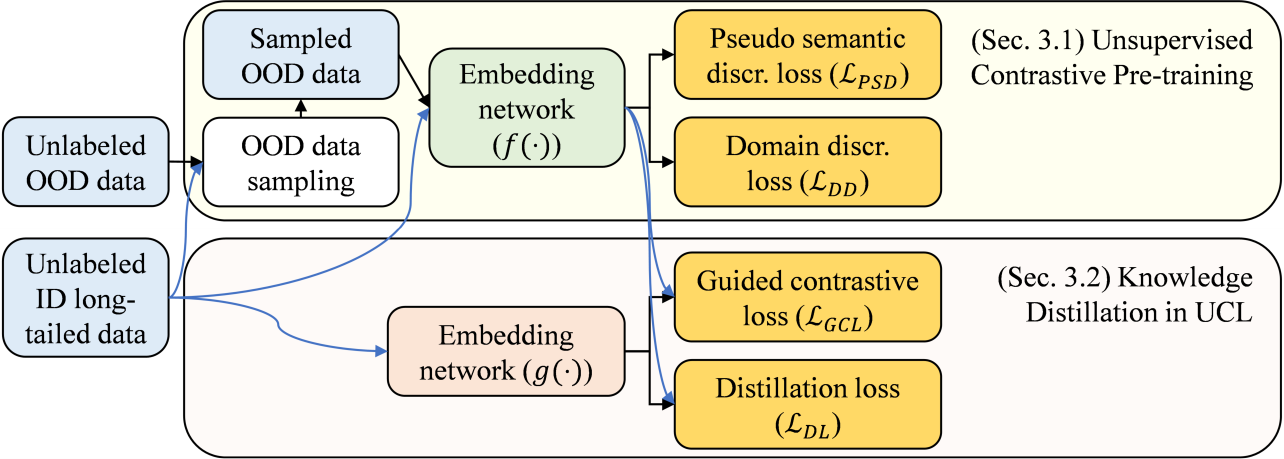}}
\end{minipage}
   \caption{Overview of the proposed framework during training. In the unsupervised contrastive pre-training stage, an embedding network $f(\cdot)$ is trained using ID long-tailed data and sampled OOD data by back-propagating pseudo semantic discrimination loss $\calL_{PSD}$ and domain discrimination loss $\calL_{DD}$. In the knowledge distillation stage, the final embedding network $g(\cdot)$ is trained using ID data and the pre-trained network $f(\cdot)$ by back-propagating guided contrastive loss $\calL_{GCL}$ and distillation loss $\calL_{DL}$.}
\label{fig:framework_training}
\end{figure*}

To learn an optimal embedding space for a long-tailed dataset without any labels, we propose to first pre-train an embedding network using sampled out-of-distribution (OOD) data and in-domain (ID) data and then further optimize the network for the ID data by leveraging the pre-trained network, as presented in~\fref{fig:framework_training}.

Because most data in a long-tailed dataset belong to head classes, embedding networks trained using typical unsupervised contrastive learning methods tend to poorly represent samples belonging to tail classes. This results in limited performance in downstream tasks. To overcome this limitation, we leverage OOD data that are close to the tail samples to rebalance the data distribution for unsupervised contrastive learning, as presented in~\sref{sec:stage1}. The rebalanced data contribute to learning a more balanced and well-separated embedding space.

However, the learned embedding space using the rebalanced data might not be optimal for the target ID data. Additionally, the sampled OOD data for the tail classes might be noisy. Hence, to obtain an optimal embedding network for the ID data, we further train the network using the ID data and the pre-trained network, as described in~\sref{sec:stage2}. The pre-trained network is utilized to select proper positive/negative samples and to control strengths of attractive/repulsive forces. Additionally, the pre-trained network is employed to transfer its balanced and well-separated embedding space to the training network.

\subsection{Unsupervised Contrastive Pre-training}
\label{sec:stage1}
We investigate unsupervised contrastive pre-training that leverages OOD data alongside long-tailed ID data to learn a more balanced embedding space. Additionally, we present a method that assigns pseudo semantic labels and encodes semantic relationships to enhance linear separability between samples of different classes. The pre-trained network is utilized to learn a further optimal embedding network for ID data in~\sref{sec:stage2}. 

\subsubsection{Sampling Out-Of-Domain Data}
To learn a balanced embedding space for a long-tailed dataset, we sample additional images that seem to contain objects of tail classes from an out-of-distribution (OOD) dataset $\calD^{OOD}$. To select OOD images that are closely related to tail classes, we first cluster the feature embeddings of in-domain (ID) images $\calD^{ID}$ into $N_c$ clusters where each cluster is interpreted as a semantic class. We then compute a tailness score for each cluster, indicating the likelihood of containing objects of a tail class. Subsequently, we allocate a sampling budget to each cluster based on its tailness score. Finally, we sample OOD images that are close to the centroids of these clusters.

\vspace{1mm}
\noindent \textbf{In-Domain Sample Clustering.}
To cluster the embeddings of long-tailed ID images, we adapt the Kullback–Leibler (KL) divergence-based clustering method in~\citep{zhang2021supporting}. While \cite{bai2023effectiveness} utilized a standard $k$-means clustering algorithm, it tends to assign most cluster centroids near samples belonging to head classes, with only a few centers close to samples of tail classes. Consequently, it has limitations in grouping samples from tail classes, which in turn results in challenges in allocating sampling budgets and sampling OOD images for tail classes.

Given the entire ID images $\calD^{ID}=\{\mI_i\}_{i=1}^{N_{ID}}$, we first apply an embedding network $f(\cdot)$ to extract their feature embeddings $\calZ^{ID}=\{\vz_i\}_{i=1}^{N_{ID}}$ where $\vz_i=f(\mI_i)$. $N_{ID}$ denotes the number of samples in $\calD^{ID}$. We then initialize cluster centroids using cluster centers obtained by applying a standard $k$-means clustering algorithm to the feature embeddings $\calZ^{ID}$. Subsequently, we refine the cluster centroids iteratively using the KL divergence-based clustering method.

To refine the centroids $\{\vmu_k\}_{k=1}^{N_c}$, we first perform a soft assignment of the feature embeddings $\vz_i$ to the clusters, where $N_c$ denotes the number of clusters. Specifically, following~\citep{van2008visualizing}, we employ Student's $t$-distribution to estimate the probability $q_{ik}$ that sample $\vx_i$ belongs to cluster $k$ based on the distance between the sample's embedding vector $\vz_i$ and the cluster center $\vmu_k$.
\begin{equation}
q_{ik} := \frac{(1+\norm{\vz_i-\vmu_k}_2^2 / d)^{-\frac{d+1}{2}}}{\sum_{\bar{k}=1}^{N_c}(1+\norm{\vz_i-\vmu_{\bar{k}}}_2^2 / d)^{-\frac{d+1}{2}}}
\label{eqn:student}
\end{equation}
where $d$ denotes the degree of freedom of the $t$-distribution. We set $d=1$ following~\citep{zhang2021supporting} due to the absence of a validation set to search for this hyperparameter.

We then iteratively refine the centroids $\{\vmu_k\}_{k=1}^{N_c}$ by minimizing the KL divergence between the current soft assignment $q_{ik}$ and the auxiliary target assignment probability, following~\citep{xie2016unsupervised}. The target distribution $p_{ik}$ is defined as follows:
\begin{equation}
p_{ik} := \frac{q^2_{ik} / h_k}{\sum_{\bar{k}=1}^{N_c} q^2_{i\bar{k}} / h_{\bar{k}}} 
\text{\hspace{0.2cm} where \hspace{0.2cm}} h_k := \sum_{i=1}^B q_{ik}. 
\label{eqn:refine}
\end{equation}
$B$ represents the number of samples in a minibatch. $h_k$ is the soft sample frequency for cluster $k$. Squaring of $q_{ik}$ emphasizes reliance on highly confident samples while reducing the influence of samples with lower confidence. $h_k$ normalizes to balance clusters with different sample counts. 

Then, the KL divergence-based loss is computed as follows:
\begin{equation}
\calL_{cluster} := \frac{1}{B} \sum_{i=1}^B \sum_{k=1}^{N_c} p_{ik} \log \frac{p_{ik}}{q_{ik}}_.
\label{eqn:kl}
\end{equation}
The iterative refinement of $\vmu_k$ using $\calL_{cluster}$ terminates when the proportion of samples with changed cluster assignments falls below a threshold. After termination, each sample $\vx_i$ is assigned to the closest cluster $k$ based on the L2 distance between $\vz_i$ and $\vmu_k$.

\vspace{1mm}
\noindent \textbf{Cluster-Wise Tailness Score Estimation.}
We compute the cluster-wise scores by first estimating the score of each ID sample belonging to tail classes and then averaging the scores of the samples within each cluster. 

To estimate the tailness score for an instance $\vx_i$, we first form a set $\calZ^i$ containing $\vz_i$ and its $K$ nearest neighbors in the embedding space given $\calZ^{ID}=\{\vz_i\}_{i=1}^{N_{ID}}$. Subsequently, we compute the cosine similarities $s_{cos}(\vz_i, \vz_j)$ between all pairs of the samples in $\calZ^i$. Since samples belonging to head classes tend to have higher densities in the embedding space than those of tail classes, we use the cosine similarities to estimate the density of each sample's neighborhood and to compute its tailness score. Specifically, the score $\hat{s}^{t}_i$ for sample $\vx_i$ is computed as follows:
\begin{equation}
\hat{s}^{t}_i := - \frac{\sum_{\vz_m \in \calZ^i} \sum_{\vz_n \in (\calZ^i \setminus \vz_m) } \exp (s_{cos}(\vz_m, \vz_n))}{K(K+1)} 
\label{eqn:tailness_score_current}
\end{equation}
where $t$ denotes the current epoch during training.

To enhance the robustness of the estimated scores, we apply momentum updates to them. Accordingly, the tailness score $s^t_i$ for sample $\vx_i$ is updated as follows:
\begin{equation}
s^t_i := \rho s^{(t-T)}_i + (1-\rho) \hat{s}^t_i 
\label{eqn:tailness_score_momentum}
\end{equation}
where $\rho$ represents the momentum hyperparameter with a value between 0 and 1. When $t=0$, $s^0_i = \hat{s}^0_i$. The score $s^t_i$ is updated every $T$ epochs for efficiency. A higher score $s^t_i$ indicates that the corresponding sample $\vx_i$ has a sparser neighborhood in the embedding space, implying a higher probability of belonging to tail classes.

Lastly, we compute the cluster-wise tailness score $s_k^{tail}$ by averaging the instance-wise scores $s^t_i$ of the samples within each cluster as follows: 
\begin{equation}
s_k^{tail} := \frac{1}{|\calC_k|} \sum_{i \in \calC_k} s^t_i
\label{eqn:cluster_tailness}
\end{equation}
where $\calC_k$ denotes the set containing the sample indices in cluster $k$, and $|\calC_k|$ represents the number of samples in $\calC_k$. While \cite{bai2023effectiveness} estimated instance-wise tailness scores using only the samples in a minibatch, we compute them using the entire data for better accuracy and robustness.

\vspace{1mm}
\noindent \textbf{Sampling Budget Allocation \& OOD Data Sampling.}
We allocate more sampling budgets to the clusters with higher cluster-wise tailness scores $s_k^{tail}$. Following~\citep{bai2023effectiveness}, the sampling budget $\vu_k$ for cluster $k$ is assigned as follows:
\begin{equation}
\vu_k := N_b \cdot S\Big(\frac{\hat{\vec{s}}^{tail}}{\tau}\Big)_k \text{\hspace{0.1cm} where \hspace{0.1cm}}
\hat{s}_k^{tail} := \frac{s_k^{tail} - \mu(\vec{s}^{tail})}{ \sigma(\vec{s}^{tail})}.
\label{eqn:sampling_budget}
\end{equation}
$N_b$ denotes the total sampling budget; $S(\cdot)$ represents a softmax function; $\hat{s}_k^{tail}$ is the normalized cluster-wise tailness score; $\mu(\cdot)$ and $\sigma(\cdot)$ denote mean and standard deviation operations, respectively; $\tau$ represents the temperature hyperparameter.

Based on the allocated sampling budget for each cluster, we sample OOD images whose feature embeddings are close to the cluster center. We denote the set containing all $N_b$ sampled OOD images as $\calD_s^{OOD}$. We update the set $\calD_s^{OOD}$ every $T$ epochs to deal with changes in the embedding space.

\subsubsection{Pre-training With Sampled OOD Data}
To enhance the separability of samples from different classes, we investigate a method that assigns pseudo semantic labels and encodes semantic relationships using both ID and sampled OOD data. Typical unsupervised contrastive learning (UCL) methods employ instance discrimination tasks to train an embedding network due to the absence of labels. Specifically, they use an augmented version of an instance as its positive sample and other instances as negative samples. Consequently, they also push apart instances from the same class. Therefore, we utilize the proximity of samples to encode semantic relationships.
Moreover, the embedding network learns a more balanced embedding space by incorporating sampled OOD data into long-tailed ID data since instances in the merged dataset are more uniformly distributed in the embedding space.

In detail, we first sample an instance $\vx_i$ from the merged dataset $\calD$ containing instances of both the ID dataset $\calD^{ID}$ and the sampled OOD dataset $\calD_s^{OOD}$. We then find $K^{pos}$ nearest neighbors based on cosine similarity in the embedding space. The nearest neighboring samples are considered as positive samples in addition to the instance's augmented sample, while the other samples in a minibatch are used as negative samples. Subsequently, we train the embedding network to increase the similarities between the instance and positive samples while reducing the similarities between the instance and negative samples. Since we consider instance-wise nearest neighbors as samples with the same pseudo semantic class, we refer to the corresponding loss as the pseudo semantic discrimination loss. The loss $\calL_{PSD}$ is computed as follows: 
\begin{equation}
\calL_{PSD} := -\frac{1}{|\calZ^{B}|} \sum_{\vz_i \in \calZ^{B}} \log \frac{ \sum_{\vz_j \in \calS_i^{pos}} \exp(\vz_i \cdot \vz_j / \tau)}{ \sum_{\vz_j \in \calS_i^{neg} } \exp(\vz_i \cdot \vz_j / \tau)} 
\label{eqn:semantic_loss}
\end{equation}
where $\calZ^{B}$ represents the set containing the embeddings of instances in the minibatch. $\calS_i^{pos}$ and $\calS_i^{neg}$ denote sets consisting of the embeddings of positive samples and negative samples of $\vz_i$, respectively. $|\calZ^{B}|$ represents the number of samples in $\calZ^{B}$. $\tau$ denotes the temperature hyperparameter. The nearest neighboring instances are selected among the samples within the instance's domain (either ID or OOD). The nearest neighbors for each instance are updated every $T$ epochs.

In addition to the pseudo semantic discrimination loss, we utilize the domain discrimination loss in~\citep{bai2023effectiveness}. Unlike typical unsupervised contrastive learning, we use a combined dataset containing samples from two different domains and know the origin domain of each sample. Accordingly, the domain discrimination loss is computed using the domain information as ground-truth labels, which aims to bring samples from the same domain closer together while pushing those from different domains farther apart. The loss $\calL_{DD}$ is calculated as follows:
\begin{equation}
\begin{split}
\calL_{DD} :=& -  \frac{1}{|\calZ^{B}|} \sum_{\vz_i \in \calZ^{B}} 
\frac{1}{|\calS_i^{s}|} \sum_{\vz_p \in \calS_i^{s}} \\ &\log 
 \bigg( \frac{\exp(\vz_i \cdot \vz_p / \tau)}{\exp(\vz_i \cdot \vz_p / \tau) + \sum_{\vz_n \in \calS_i^{d} } \exp(\vz_i \cdot \vz_n / \tau)} \bigg)
\end{split}
\label{eqn:domain_discrimination}
\end{equation}
where $\calZ^{B}$ represents the set containing the embeddings of instances in the minibatch. $\calS_i^{s}$ and $\calS_i^{d}$ denote sets consisting of embeddings of samples within the same domain and those from the different domain compared to $\vz_i$, respectively. $|\calS_i^{s}|$ represents the number of samples in $\calS_i^{s}$. 

The total contrastive pre-training loss $\calL_{CPT}$ is computed through a weighted summation of $\calL_{PSD}$ and $\calL_{DD}$ as follows:
\begin{equation}
\calL_{CPT} := \calL_{PSD} + \alpha \calL_{DD}
\label{eqn:contrastive_loss}
\end{equation}
where $\alpha$ is a hyperparameter used to balance the two losses.

\begin{figure*}[!h] 
\centerline{\includegraphics[scale=0.7]{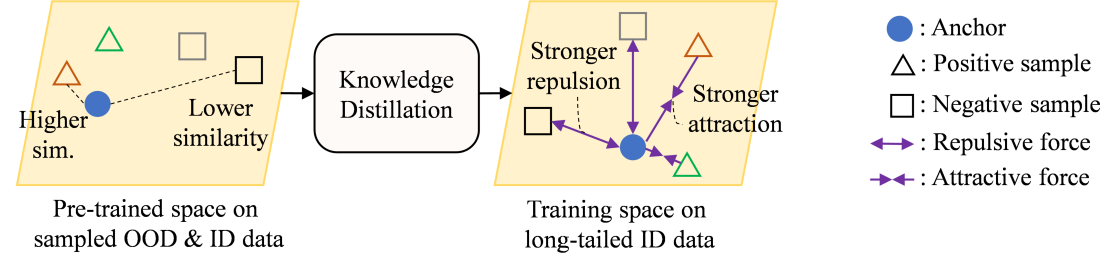}}
   \caption{Proposed method leverages the balanced and well-separated embedding space of a pre-trained network to train a further optimal network for in-domain data. The pre-trained network is trained on out-of-distribution data and in-domain data.}   
\label{fig:teaser}
\end{figure*}

\subsection{Knowledge Distillation in UCL}
\label{sec:stage2}
The pre-trained embedding network in~\sref{sec:stage1} encodes balanced embeddings with good separability between samples of different classes by incorporating sampled OOD data and learning semantic and domain discriminations. However, the pre-trained network might not be optimal for the ID dataset. Therefore, we train a further optimal embedding network for the ID dataset while using the pre-trained network as a guiding network in unsupervised contrastive learning (UCL). We leverage the balanced and well-separated embedding space from the pre-trained network to select valid positive/negative samples and to control strengths of attractive/repulsive forces in contrastive learning, as presented in~\fref{fig:teaser}. Additionally, we distil the embedding space from the pre-trained network and transfer it to the training network.

In detail, we first initialize a new embedding network $g(\cdot)$ using the pre-trained and frozen guiding network $f(\cdot)$. Subsequently, we further optimize the network $g(\cdot)$ using the ID dataset $\calD^{ID}$ and the guiding network $f(\cdot)$. To train $g(\cdot)$, we sample instances from $\calD^{ID}$ and extract their embeddings using the guiding network $f(\cdot)$. We then utilize the embeddings to select positive and negative samples since the guiding network is already capable of semantic discrimination. For each instance, a positive sample is randomly selected from the $K^{kd}$ nearest neighbors in the embedding space, and a negative sample is randomly chosen among instances in the farthest cluster from the instance's cluster. A cluster index is assigned to each ID data by applying the clustering method in~\sref{sec:stage1} to the embeddings from the frozen network $f(\cdot)$.\footnote{Clustering is performed only once at the beginning due to the fixed embeddings from the frozen network.} Finally, we train the new network $g(\cdot)$ using the sampled instances and their positive/negative samples by minimizing the proposed guided contrastive loss and the distillation loss.

For the guided contrastive loss, given an instance $\vx_i$ and its positive/negative samples ($\vx_i^{p}$, $\vx_i^{n}$), we first extract embeddings ($\vz_i$, $\vz_i^{p}$, $\vz_i^{n}$) using the guiding network $f(\cdot)$ and embeddings ($\vy_i$, $\vy_i^{p}$, $\vy_i^{n}$) using the current embedding network $g(\cdot)$. Then, we compute the cosine similarity $w_i^{pos}$ between $\vz_i$ and $\vz_i^{p}$ and the similarity $w_i^{neg}$ between $\vz_i$ and $\vz_i^{n}$. Subsequently, these similarities ($w^{pos}$, $w^{neg}$) are utilized as weighting coefficients in the guided contrastive loss. The coefficients control the strengths of attractive/repulsive forces in contrastive learning. Positive samples with higher $w_i^{pos}$ are heavily pulled closer together, while negative samples with lower $w_i^{neg}$ are strongly pushed farther apart. Consequently, the guided contrastive loss $\calL_{GCL}$ is computed as follows:
\begin{equation}
\begin{split}
\calL_{GCL} := \frac{1}{|\calX^{B}|} \sum_{i \in \calX^{B}} \Big[(1+w_i^{pos})(1 - s_{cos}(\vy_i, \vy_i^p)) \\ 
+ (1 - w_i^{neg}) (1+s_{cos}(\vy_i, \vy_i^n)) \Big]
\end{split}
\label{eqn:guided_contrastive} 
\end{equation}
where $\calX^{B}$ represents the set containing the indices of instances in the minibatch. $s_{cos}(\cdot,\cdot)$ denotes the cosine similarity function. $|\calX^{B}|$ represents the number of samples in $\calX^{B}$.

In addition to the guided contrastive loss, we employ a distillation loss to maintain the embedding structure of the guiding network in the new embedding network. The distillation loss is computed by comparing the cosine similarities from the guiding network $f(\cdot)$ to those from the current network $g(\cdot)$. Consequently, the loss is calculated as follows:
\begin{equation}
\calL_{DL} := \frac{1}{|\calX^{B}|(|\calX^{B}|-1)} \sum_{i \in \calX^{B}} \sum_{j \in (\calX^{B} \setminus i)} 
 \Big[ \big(s_{cos}(\vz_i, \vz_j) - s_{cos}(\vy_i, \vy_j) \big)^2   \Big]
\label{eqn:distillation} 
\end{equation}
where $\calX^{B}$ represents the set containing the indices of instances in the minibatch.

The total guided learning loss is computed through a weighted summation of $\calL_{GCL}$ and $\calL_{DL}$ as follows:
\begin{equation}
\calL_{GL} := \calL_{GCL} + \beta \calL_{DL}
\label{eqn:guided_loss}
\end{equation}
where $\beta$ is a hyperparameter used to balance the two losses.

\begin{table*}[!h]
\centering
\caption{Quantitative comparison of accuracies and balancedness using the CIFAR-10-LT and CIFAR-100-LT datasets~\citep{cui2019class}.}
\label{tab:result_cifar}
\begin{minipage}{1\linewidth}
\centering
\begin{tabular}{>{\centering}m{0.1\textwidth}|>{}m{0.3\textwidth}|>{\centering}m{0.08\textwidth} >{\centering}m{0.08\textwidth}  *{2}{>{\centering}m{0.08\textwidth}} >{\centering\arraybackslash}m{0.08\textwidth}} 
\toprule
Dataset & \centering{Method} & Many $\uparrow$ & Med. $\uparrow$ & Few $\uparrow$ & STD $\downarrow$ & All $\uparrow$  \\
\midrule
 & SimCLR~\citep{chen2020simple} & 82.40 & 73.91 & 70.19 & 5.11 & 75.34  \\
 & SimCLR + COLT~\citep{bai2023effectiveness} & 87.50 & 81.65 & 80.80 & 2.98 & 83.15  \\
CIFAR & SimCLR + KDUP (ours) &\textbf{93.80} &\textbf{89.17} &\textbf{87.36} &\textbf{2.71} &\textbf{90.26} \\
\cmidrule{2-7} 
-10-LT & SDCLR~\citep{jiang2021self} & 86.69 & 82.15 & 76.23 & 4.28 & 81.74 \\
& SDCLR + COLT~\citep{bai2023effectiveness} & 90.87 & 84.28 & 81.45 & 3.95 & 85.41 \\
& SDCLR + KDUP (ours)  &\textbf{95.23} &\textbf{89.79} &\textbf{88.65} &\textbf{2.87} &\textbf{91.42}  \\
\midrule
 & SimCLR~\citep{chen2020simple} &51.50 &45.58 &45.96&2.71 &47.65  \\
 & SimCLR + COLT~\citep{bai2023effectiveness} &57.94 &56.74 &57.72  &\textbf{0.52} &57.46 \\
CIFAR & SimCLR + KDUP (ours) &\textbf{63.83} &\textbf{62.47} &\textbf{61.96} & 0.79 &\textbf{62.36} \\
\cmidrule{2-7} 
-100-LT & SDCLR~\citep{jiang2021self} &58.54 &55.70 &52.10 &2.64  &55.48  \\
& SDCLR + COLT~\citep{bai2023effectiveness} &63.28 & 60.85 &59.42 &1.59 &61.18  \\
& SDCLR + KDUP (ours) &\textbf{66.95}  &\textbf{64.56} &\textbf{63.45} &\textbf{1.46} & \textbf{64.98} \\
\bottomrule 
\end{tabular}
\end{minipage}
\end{table*}

\section{Experiments and Results}
\subsection{Experimental Setting}
\label{sec:experimental_setting}
\noindent \textbf{Dataset.} 
Experiments are conducted using four public long-tailed datasets: CIFAR-10-LT, CIFAR-100-LT, ImageNet-100-LT, and Places-LT, following previous work~\citep{bai2023effectiveness}. CIFAR-10-LT and CIFAR-100-LT datasets~\citep{cui2019class} contain subsets of images from the original CIFAR-10 and CIFAR-100 datasets~\citep{krizhevsky2009learning}, respectively. The imbalance ratio of 100 is used. We utilize 300K Random Images from~\citep{hendrycks2018deep} as the OOD dataset, following~\citep{wei2022open}. ImageNet-100-LT~\citep{jiang2021self} comprises 12K images sampled from ImageNet-100~\citep{tian2020contrastive} using a Pareto distribution. The number of instances for each class ranges from 5 to 1,280. ImageNet-R~\citep{hendrycks2021many} is used as the OOD dataset. Places-LT~\citep{liu2019large} consists of about 62.5K images sampled from Places dataset~\citep{zhou2017places} using a Pareto distribution. The number of instances for each class ranges from 5 to 4,980. We employ Places-Extra69~\citep{zhou2017places} as the OOD dataset.

\vspace{1mm}
\noindent \textbf{Evaluation Protocol.} 
We employ two standard evaluation protocols: linear probing and few-shot learning. Initially, we train the embedding model $g(\cdot)$ with the proposed method using unlabeled data. Subsequently, we freeze the model $g(\cdot)$ and append a linear classifier to the frozen model. Finally, we train only the linear classifier, using the entire data with labels for linear probing and 1\% of the labeled data for few-shot learning, respectively.

After training the linear classifier, we compute accuracies to evaluate the separability and balancedness of the learned embedding space. The overall accuracy is utilized to assess the linear separability. To measure its balancedness, we divide a test dataset into three disjoint groups (\ie Many, Medium, Few) based on the number of instances for each class. Then, we assess the balancedness by calculating the standard deviation of the accuracies for these three groups.

\vspace{1mm}
\noindent \textbf{Implementation Details.} We apply the proposed method to SimCLR~\citep{chen2020simple} and SDCLR~\citep{jiang2021self} using ResNet-18 as the backbone for CIFAR-10-LT and CIFAR-100-LT datasets, following~\citep{bai2023effectiveness}. For ImageNet-100-LT and Places-LT datasets, the proposed method is applied to SimCLR~\citep{chen2020simple} with ResNet-50 as the backbone. 

We pre-train the networks for 2,000, 2,000, 1,000, and 500 epochs for the CIFAR-10-LT, CIFAR-100-LT, ImageNet-100-LT, and Places-LT datasets, respectively, using the method in~\sref{sec:stage1}. We then further optimize the networks for 500 epochs for all datasets using the method in~\sref{sec:stage2}. Lastly, for evaluation, we train the linear classifier for 30 epochs and 100 epochs for linear probing and few-shot learning protocols, respectively, for all datasets. $T$ is set to 50 epochs for the ImageNet-100-LT dataset and 25 epochs for all the other datasets. In~\sref{sec:stage1}, $K=10$, $\rho=0.9$, $N_b=10,000$, $K^{pos}=3$, and $\alpha=0.3$. In~\sref{sec:stage2}, $K^{kd}=5$ and $\beta=0.4$. The code will be made available on GitHub upon publication to ensure better reproducibility.

\subsection{Result}
We present quantitative results of the proposed method and previous works on the CIFAR-10-LT and CIFAR-100-LT datasets~\citep{cui2019class} in~\tref{tab:result_cifar}. Following~\citep{bai2023effectiveness}, linear probing is utilized for evaluation. For the CIFAR-10-LT dataset, the proposed method outperforms COLT~\citep{bai2023effectiveness} by 7.11\% and 6.01\% in overall accuracy with 0.27 and 1.08 lower standard deviation (STD) when integrated with SimCLR~\citep{chen2020simple} and SDCLR~\citep{jiang2021self}, respectively. For the CIFAR-100-LT dataset, the proposed framework achieves 4.9\% and 3.8\% higher overall accuracies with 0.27 higher and 0.13 lower STD compared to COLT~\citep{bai2023effectiveness} with SimCLR~\citep{chen2020simple} and SDCLR~\citep{jiang2021self}, respectively. While the higher STD of the proposed method with SimCLR represents lower balancedness compared with the previous work, it is worth noting that our absolute STD value is only 0.79, which is still very low.

\tref{tab:result_imagenet} shows quantitative comparisons of the proposed method with previous works on the ImageNet-100-LT~\citep{jiang2021self} and Places-LT~\citep{liu2019large} datasets. All methods in this table are implemented on top of SimCLR~\citep{chen2020simple} and evaluated using both linear probing and few-shot learning protocols. The proposed method consistently outperforms the previous state-of-the-art method for all datasets and protocols by achieving higher accuracies and lower STDs. For the ImageNet-100-LT dataset, the proposed method achieves 1.61\% and 2.19\% higher overall accuracies with 0.12 and 0.19 lower STDs than SimCLR+COLT~\citep{bai2023effectiveness} in few-shot learning and linear probing, respectively. For the Places-LT dataset, the proposed framework outperforms SimCLR+COLT~\citep{bai2023effectiveness} by 0.76\% and 1.45\% in overall accuracies with 0.11 and 0.22 lower STDs in few-shot learning and linear probing, respectively.

\begin{table*}[!h]
\centering
\caption{Quantitative comparison of accuracies and balancedness using the ImageNet-100-LT~\citep{jiang2021self} and Places-LT datasets~\citep{liu2019large}.}
\label{tab:result_imagenet}
\begin{minipage}{1\linewidth}
\centering
\begin{tabular}{>{\centering}m{0.08\textwidth}|>{\centering}m{0.08\textwidth}|>{\centering}m{0.23\textwidth}|>{\centering}m{0.08\textwidth}>{\centering}m{0.08\textwidth} *{2}{>{\centering}m{0.08\textwidth}} >{\centering\arraybackslash}m{0.08\textwidth}} 
\toprule
Dataset & Protocol & Method & Many $\uparrow$ & Med. $\uparrow$ & Few $\uparrow$ & STD $\downarrow$ & All $\uparrow$  \\
\midrule
 & & Random sampling &  52.82 &  42.88 &  40.27 &  5.41 & 46.42 \\
 & Few & MAK~\citep{jiang2021improving} &  54.33 &  45.01 &  40.12 &  5.90 &  48.01 \\
 & -shot & COLT~\citep{bai2023effectiveness} &  54.26 &  46.54 &  43.38 &  4.57 &  49.14 \\
ImageNet & & KDUP (ours) &  \textbf{56.08} & \textbf{48.41}  &  \textbf{45.59} &  \textbf{4.45} &  \textbf{50.75} \\
\cmidrule{2-8} 
-100-LT & & Random sampling & 74.33 & 68.52 & 62.65 & 4.77 & 70.02 \\
 & Linear & MAK~\citep{jiang2021improving} &  75.57 &  68.20 &  66.29 &  4.07  &  70.83 \\
 & -probing & COLT~\citep{bai2023effectiveness} & 75.13 & 71.38 &  66.62  &  3.48 &  72.22 \\
 & & KDUP (ours) &  \textbf{77.91} &  \textbf{74.07} &  \textbf{69.84} &  \textbf{3.29} &  \textbf{74.41} \\
\midrule
 & &  Random sampling &  30.61 & 33.94 &  37.55 &  2.83 &   33.45\\
 & Few- &  MAK~\citep{jiang2021improving} &  30.41& 34.47& 37.59& 2.94& 33.62 \\
 & shot &  COLT~\citep{bai2023effectiveness} &  31.04& 34.65 &37.49 &2.64& 33.91 \\
Places &  &  KDUP (ours) &  \textbf{31.85} &  \textbf{35.31} &  \textbf{38.02} &  \textbf{2.53} &  \textbf{34.67} \\
\cmidrule{2-8} 
-LT &   &  Random sampling &  40.21 &  47.59 &  50.51  &  4.33 &  45.51 \\
 & Linear  & MAK~\citep{jiang2021improving} &  40.83 &47.78 &50.72& 4.15& 45.86 \\
 & -probing & COLT~\citep{bai2023effectiveness} &  41.55& 48.40& 50.54 &3.83& 46.36 \\
 &   & KDUP (ours) &  \textbf{43.18} &  \textbf{49.74} & \textbf{51.61} & \textbf{3.61} &  \textbf{47.81} \\
\bottomrule 
\end{tabular}
\end{minipage}
\end{table*}

\begin{figure*}[!h]
\begin{minipage}{0.325\linewidth}  
\centerline{\includegraphics[scale=0.26]{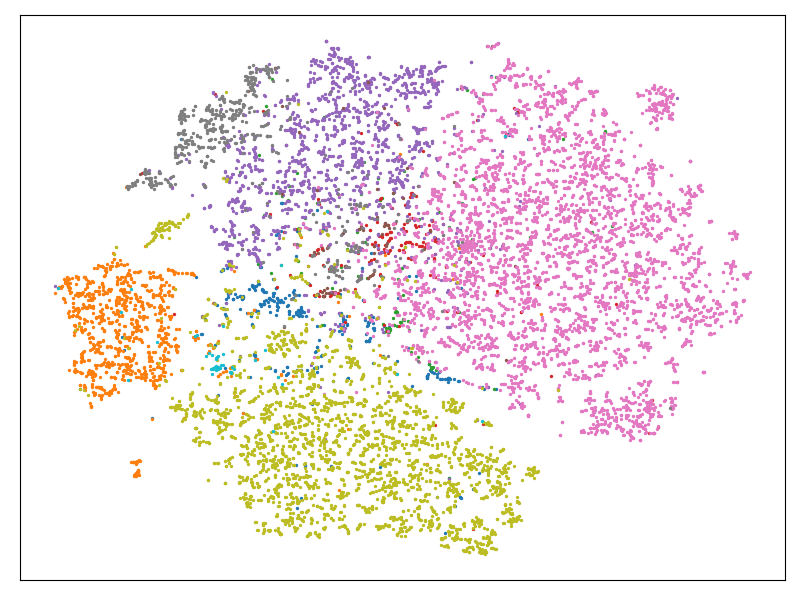}}
\end{minipage}
\begin{minipage}{0.325\linewidth}  
\centerline{\includegraphics[scale=0.26]{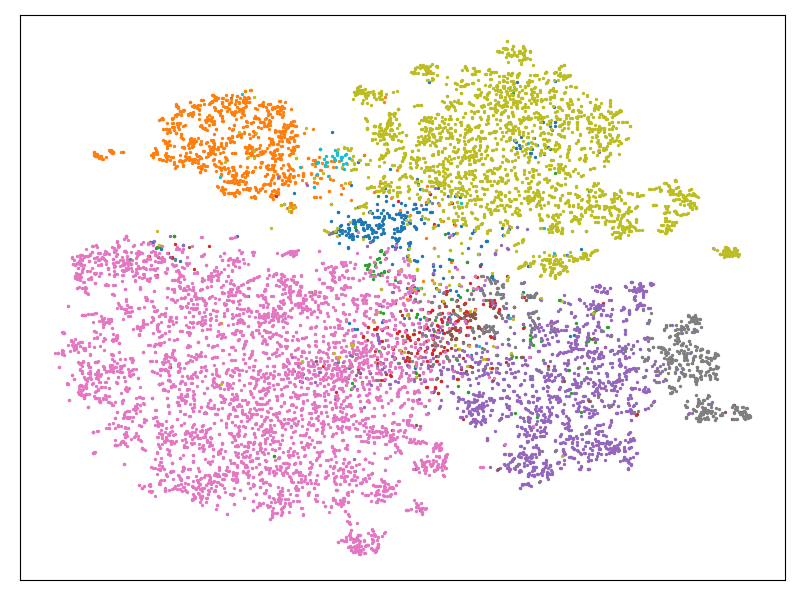}}
\end{minipage}
\begin{minipage}{0.325\linewidth}  
\centerline{\includegraphics[scale=0.26]{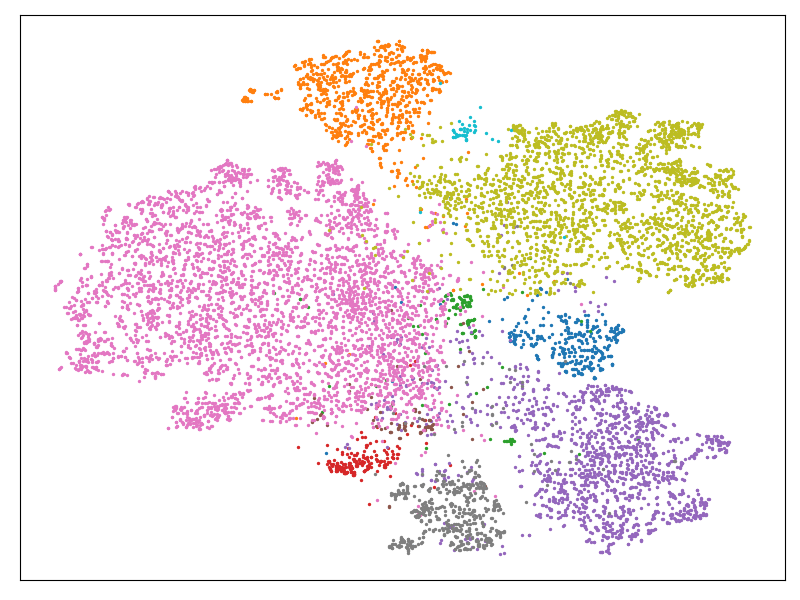}}
\end{minipage}
\\
\begin{minipage}{0.325\linewidth}
\centerline{SimCLR~\citep{chen2020simple}}
\end{minipage}
\begin{minipage}{0.325\linewidth}
\centerline{SimCLR+COLT~\citep{bai2023effectiveness}}
\end{minipage}
\begin{minipage}{0.325\linewidth}
\centerline{SimCLR+KDUP (ours)}
\end{minipage}
\caption{t-SNE visualization of the CIFAR-10-LT dataset~\citep{cui2019class}.}
\label{fig:result}
\end{figure*}

\fref{fig:result} visualizes a t-SNE projection of the learned representations of the data in the CIFAR-10-LT dataset~\citep{cui2019class}. This figure demonstrates that the proposed method has better separability between samples of different classes and forms more compact clusters for each class than previous works~\citep{chen2020simple,bai2023effectiveness}.

Additionally, in~\tref{tab:result_CHI_DBI}, we quantitatively compare the learned representations of the proposed method with those of the previous state-of-the-art method, COLT~\citep{bai2023effectiveness}, using the Calinski-Harabasz Index (CHI)~\citep{CHI1974} and the Davies-Bouldin Index (DBI)~\citep{DBI1979}. All the scores in the table are computed using the models based on SimCLR~\citep{chen2020simple}. In addition to the results shown in~\fref{fig:result}, the measured CHI and DBI scores demonstrate that the proposed method forms more compact clusters for each class while achieving better separation between different classes than the previous work.

\begin{table*}[!t]
\centering
\caption{Quantitative comparison of the learned representations using the Calinski-Harabasz Index (CHI) and the Davies-Bouldin Index (DBI) on the CIFAR-10-LT, CIFAR-100-LT, ImageNet-100-LT, and Places-LT datasets.}
\label{tab:result_CHI_DBI}
\begin{tabular}{>{\centering}m{0.2\textwidth}|>{}m{0.25\textwidth}|>{\centering}m{0.15\textwidth} >{\centering\arraybackslash}m{0.15\textwidth}}
\toprule
Dataset & \centering{Method} & CHI $\uparrow$ & DBI $\downarrow$ \\ 
\midrule
\multirow{2}{*}{CIFAR-10-LT} & COLT~\citep{bai2023effectiveness} &143.62 & 3.38   \\
 					& KDUP (ours) & \textbf{187.03} & \textbf{2.21}  \\
\midrule
\multirow{2}{*}{CIFAR-100-LT} & COLT~\citep{bai2023effectiveness} &103.47 & 4.25   \\
 					& KDUP (ours) & \textbf{132.58} & \textbf{3.64}  \\
\midrule
\multirow{2}{*}{ImageNet-100-LT} & COLT~\citep{bai2023effectiveness} &121.63 & 3.79   \\
 					& KDUP (ours) & \textbf{155.74} & \textbf{2.85}  \\
\midrule
\multirow{2}{*}{Places-LT} & COLT~\citep{bai2023effectiveness} & 89.82 & 5.83   \\
 					& KDUP (ours) & \textbf{102.71} & \textbf{4.97}  \\
\bottomrule
\end{tabular}
\end{table*}

\subsection{Analysis}
We present an ablation study of the proposed framework using the CIFAR-10-LT dataset~\citep{cui2019class} in~\tref{tab:ablation}. The experiments are conducted with models based on SimCLR~\citep{chen2020simple} and the linear probing protocol. 
Firstly, we construct a strong baseline model by incorporating the KL divergence-based clustering method instead of $k$-means clustering and the proposed tailness score estimation. The baseline model achieves 1.56\% higher accuracy and 0.10 lower STD than SimCLR+COLT~\citep{bai2023effectiveness}. By employing the proposed pseudo semantic discrimination method, we further enhance overall accuracy by 3.74\% with 0.08 lower STD. Finally, the proposed knowledge distillation method in~\sref{sec:stage2} further improves overall accuracy by 1.81\% with 0.09 lower STD. Additionally, we present the results of the strong baseline model excluding the domain discrimination loss $\calL_{DD}$\footnote{Please note that COLT~\citep{bai2023effectiveness} also employs the domain discrimination loss $\calL_{DD}$.}.


\begin{table*}[!h]
\centering
\caption{Ablation study of the proposed method on the CIFAR-10-LT dataset~\citep{cui2019class}.}
\label{tab:ablation}
\begin{minipage}{1\linewidth}
\centering
\begin{tabular}{>{}m{0.32\textwidth} | *{2}{>{\centering}m{0.09\textwidth}} *{2}{>{\centering}m{0.08\textwidth}} >{\centering\arraybackslash}m{0.08\textwidth}}
\toprule
\centering{Method} & Many $\uparrow$ & Medium $\uparrow$ & Few $\uparrow$ & STD $\downarrow$ & All $\uparrow$  \\
\midrule
SimCLR~\citep{chen2020simple} & 82.40 & 73.91 & 70.19 & 5.11 & 75.34  \\
\midrule
Strong Baseline w/o $\calL_{DD}$ & 87.31 & 81.79 & 80.24 & 3.67 & 83.25 \\
Strong Baseline 								  & 88.93 & 83.07 & 82.59 & 2.88 & 84.71\\
+ Semantic Discrimination $\calL_{PSD}$ & 92.06 & 87.82 & 85.27 & 2.80  & 88.45 \\
+ Knowledge Distillation $\calL_{GL}$ & \textbf{93.80} & \textbf{89.17} & \textbf{87.36} & \textbf{2.71} & \textbf{90.26}\\
\bottomrule 
\end{tabular}
\end{minipage}
\end{table*}

\begin{table*}[!t]
\centering
\caption{Analysis of the proposed guided contrastive loss $\calL_{GCL}$.}
\label{tab:analysis_guided_contrastive_loss}
\begin{minipage}{1\linewidth}
\centering
\begin{tabular}{>{}m{0.25\textwidth} | *{4}{>{\centering}m{0.1\textwidth}} >{\centering\arraybackslash}m{0.1\textwidth}}
\toprule
\centering{Method} & Many $\uparrow$ & Medium $\uparrow$ & Few $\uparrow$ & STD $\downarrow$ & All $\uparrow$  \\
\midrule
Ours w/o KD (Sec. 3.2) & 92.06 & 87.82 & 85.27 & 2.80  & 88.45 \\
+ KD with typical loss & 87.65 & 81.76 & 80.93 & 2.99 & 83.32 \\
+ Guided Sampling & 92.81 & 88.67 & 86.05 & 2.78 & 89.19 \\
+ Guided Weighting & \textbf{93.80} & \textbf{89.17} & \textbf{87.36} & \textbf{2.71} & \textbf{90.26} \\
\bottomrule 
\end{tabular}
\end{minipage}
\end{table*}

\begin{table*}[!t]
\centering
\caption{Effectiveness of the proposed method under the experimental setting of MAK~\citep{jiang2021improving}.}
\label{tab:analysis_MAK}
\begin{minipage}{1\linewidth}
\centering
\begin{tabular}{>{}m{0.3\textwidth} | *{4}{>{\centering}m{0.09\textwidth}} >{\centering\arraybackslash}m{0.09\textwidth}}
\toprule
\centering{Method} & Many $\uparrow$ & Med. $\uparrow$ & Few $\uparrow$ & STD $\downarrow$ & All $\uparrow$  \\
\midrule
MAK~\citep{jiang2021improving} & 74.7$\pm$0.2 & 69.2$\pm$0.7 & 66.6$\pm$0.7 & 3.3$\pm$0.3 & 71.1$\pm$0.5 \\
+ Semantic Discrimination $\calL_{PSD}$ & 75.8$\pm$0.4& 70.9$\pm$0.5 & 68.5$\pm$0.8 &3.0$\pm$0.5 &72.8$\pm$0.4 \\
+ Knowledge Distillation $\calL_{GL}$ & \textbf{76.3$\pm$0.3} & \textbf{72.1$\pm$0.5} & \textbf{69.1$\pm$0.6} & \textbf{2.9$\pm$0.4}  & \textbf{73.4$\pm$0.3} \\
\bottomrule 
\end{tabular}
\end{minipage}
\end{table*}

\tref{tab:analysis_guided_contrastive_loss} presents a detailed analysis of the guided contrastive loss $\calL_{GCL}$ using the models based on SimCLR~\citep{chen2020simple}, the linear probing protocol, and the CIFAR-10-LT dataset~\citep{cui2019class}. The top row shows the results of our method without the knowledge distillation (KD) stage in~\sref{sec:stage2}. The second row presents the results of employing a KD stage that uses a typical contrastive loss and $\calL_{DL}$. Because the typical contrastive loss utilizes a random sample within a mini-batch as a negative sample, this naive implementation deteriorates the overall accuracy by 5.13\%. Utilizing the pre-trained network for selecting positive/negative samples enhances the accuracy by 5.87\% and 0.74\% compared to the second and the first rows, respectively. Subsequently, controlling the strengths of attractive/repulsive forces further improves the accuracy by 1.04\%.

Additionally, \tref{tab:analysis_MAK} demonstrates the effectiveness of the proposed method using the experimental setting of~\citep{jiang2021improving}, which assumes the presence of both ID and OOD data in an external dataset. Following~\citep{jiang2021improving}, we experiment on the ImageNet-100-LT dataset while using the ImageNet-Places-Mix (IPM) as the external dataset. The IPM dataset utilizes 200 randomly sampled classes from the ImageNet-900 dataset for ID data and the Places dataset for OOD data. The experiments are conducted with models based on MAK~\citep{jiang2021improving} and the linear probing protocol. Firstly, by incorporating the proposed pseudo semantic discrimination into MAK~\citep{jiang2021improving}, the overall accuracy is improved by 1.7\% with 0.3 lower STD. The proposed knowledge distillation method in~\sref{sec:stage2} further enhances overall accuracy by 0.6\% with 0.1 lower STD. These experimental results verify that the proposed method is also effective in scenarios where an external dataset contains both ID and OOD data.

\begin{table*}[!t]
\centering
\caption{Analysis of the hyperparameters in $\calL_{CPT}$ and $\calL_{GL}$ using our method with SimCLR~\citep{chen2020simple} on the CIFAR-10-LT dataset~\citep{cui2019class}.}
\label{tab:analysis_hyperparameter}
\begin{minipage}{1\linewidth}
\centering
\begin{tabular}{>{\centering}m{0.1\textwidth}|>{\centering}m{0.1\textwidth}|>{\centering}m{0.1\textwidth} >{\centering}m{0.1\textwidth} >{\centering}m{0.1\textwidth} >{\centering}m{0.1\textwidth} >{\centering\arraybackslash}m{0.1\textwidth}} 
\toprule
$\alpha$ & $\beta$ & Many $\uparrow$ & Medium $\uparrow$ & Few $\uparrow$ & STD $\downarrow$ & All $\uparrow$  \\
\midrule
0.2 & 0.3 & 92.41 & 86.95 & 85.92 & 2.84 & 88.53  \\
0.2 & 0.4 & 92.67 & 87.96 & 86.03 & 2.79 & 88.98  \\
0.2 & 0.5 & 91.16 & 86.07 & 84.41 & 2.87 & 87.31  \\
0.3 & 0.3 & 93.21 & 88.54 & 86.65 & 2.76 & 89.54  \\
0.3 & 0.4 & \textbf{93.80} &\textbf{89.17} &\textbf{87.36} &\textbf{2.71} &\textbf{90.26}  \\
0.3 & 0.5 & 92.79 & 88.13 & 86.27 & 2.74 & 89.15  \\
0.4 & 0.3 & 92.63 & 87.54 & 86.12 & 2.79 & 88.87  \\
0.4 & 0.4 & 92.68 & 88.05 & 86.13 & 2.75 & 89.01  \\
0.4 & 0.5 & 91.63 & 86.72 & 84.93 & 2.83 & 87.84  \\
\bottomrule 
\end{tabular}
\end{minipage}
\end{table*}

\tref{tab:analysis_hyperparameter} presents an analysis of the hyperparameters in the loss functions $\calL_{CPT}$ and $\calL_{GL}$ using the proposed method based on SimCLR~\citep{chen2020simple} and the CIFAR-10-LT dataset~\citep{cui2019class}. The results indicate that the proposed method consistently outperforms previous works~\citep{chen2020simple, jiang2021self, bai2023effectiveness}, even with suboptimal hyperparameter values.

We present the effects of varying $N_b$ and $T$ in Tables~\ref{tab:analysis_sampling_budget} and~\ref{tab:analysis_sampling_frequency}, respectively. The results demonstrate that our method based on SimCLR~\citep{chen2020simple} consistently outperforms the previous SOTA method, SimCLR~\citep{chen2020simple} + COLT~\citep{bai2023effectiveness}, which achieves 83.15 with $N_b=10$K and $T=25$.

\begin{table}[!t]
\centering
\caption{Analysis of total OOD data sampling budget $N_b$ using the CIFAR-10-LT dataset~\citep{cui2019class}.}
\label{tab:analysis_sampling_budget}
\begin{minipage}{1\linewidth}
\centering
\begin{tabular}{>{\centering}m{0.13\textwidth}|>{\centering}m{0.13\textwidth}|>{\centering}m{0.13\textwidth} |>{\centering}m{0.13\textwidth} |>{\centering\arraybackslash}m{0.13\textwidth}} 
\toprule
$N_b$ & 5K & 10K & 15K & 20K  \\
\midrule
All $\uparrow$ & 88.13 & \textbf{90.26} & 89.41 & 87.65 \\
\bottomrule 
\end{tabular}
\end{minipage}
\end{table}

\begin{table}[!t]
\centering
\caption{Analysis of OOD data sampling frequency $T$ using the CIFAR-10-LT dataset~\citep{cui2019class}.}
\label{tab:analysis_sampling_frequency}
\begin{minipage}{1\linewidth}
\centering
\begin{tabular}{>{\centering}m{0.13\textwidth}|>{\centering}m{0.13\textwidth}|>{\centering}m{0.13\textwidth} |>{\centering}m{0.13\textwidth} |>{\centering\arraybackslash}m{0.13\textwidth}} 
\toprule
$T$ & 10 & 25 & 50 & 100  \\
\midrule
All $\uparrow$ & 89.25 & \textbf{90.26} & 88.73 & 87.41 \\
\bottomrule 
\end{tabular}
\end{minipage}
\end{table}

In \tref{tab:ablation_momentum}, we compare the proposed method with its variant that excludes the momentum updates for instance-wise tailness scores in~\eref{eqn:tailness_score_momentum}. The experimental results demonstrate the effectiveness of the momentum updates in tailness score estimation, especially for tail classes. We believe this is because samples belonging to tail classes are sparser than those of head classes in the embedding space, resulting in noisier tailness score estimates for tail samples. Accordingly, the momentum updates lead to greater improvements for tail classes.
\begin{table*}[!h]
\centering
\caption{Ablation study on the effectiveness of the momentum updates (MU) in~\eref{eqn:tailness_score_momentum} using the CIFAR-10-LT dataset~\citep{cui2019class}.}
\label{tab:ablation_momentum}
\begin{minipage}{1\linewidth}
\centering
\begin{tabular}{>{\centering}m{0.16\textwidth} | *{2}{>{\centering}m{0.1\textwidth}} *{2}{>{\centering}m{0.1\textwidth}} >{\centering\arraybackslash}m{0.1\textwidth}}
\toprule
Method & Many $\uparrow$ & Medium $\uparrow$ & Few $\uparrow$ & STD $\downarrow$ & All $\uparrow$  \\
\midrule
without MU & 93.04 & 89.97 & 85.74 & 2.94 & 89.13\\
with MU      & \textbf{93.80} & \textbf{89.17} & \textbf{87.36} & \textbf{2.71} & \textbf{90.26}\\
\bottomrule 
\end{tabular}
\end{minipage}
\end{table*}

In~\tref{tab:augmentedVSOOD}, we compare the results of the proposed method with those obtained using augmented data in place of OOD data. For the CIFAR-10-LT and CIFAR-100-LT datasets, the proposed method samples images from the 300K Random Images dataset for each cluster, based on the sampling budget described in Section 3.1.1. In contrast, the augmentation-based method balances the number of images in each cluster by synthesizing images through typical data augmentation techniques. Specifically, three operations are randomly selected from the following five techniques: horizontal flip, crop, rotation, color jitter, and Gaussian blur. The experimental results demonstrate the advantages of using OOD data compared to in-domain augmented images.
\begin{table*}[!h]
\centering
\caption{Comparison of using augmented data and OOD data for constructing $\calD_s^{OOD}$ on the CIFAR-10-LT and CIFAR-100-LT datasets~\citep{cui2019class}.}
\label{tab:augmentedVSOOD}
\begin{tabular}{>{\centering}m{0.15\textwidth}|>{\centering}m{0.1\textwidth}| *{4}{>{\centering}m{0.08\textwidth}} >{\centering\arraybackslash}m{0.08\textwidth}} 
\toprule
Dataset & $\calD_s^{OOD}$ & Many $\uparrow$ & Medium $\uparrow$ & Few $\uparrow$ & STD $\downarrow$ & All $\uparrow$  \\
\midrule
 \multirow{2}{*}{\centering CIFAR-10-LT} & augmented & 89.74 & 86.32 & 82.05 & 3.14 & 86.18  \\
 & OOD & \textbf{93.80}& \textbf{89.17}& \textbf{87.36}&\textbf{2.71}& \textbf{90.26}   \\
\midrule
\multirow{2}{*}{\centering CIFAR-100-LT} & augmented &61.03 &59.81 &58.37& 1.12 & 59.74  \\
 & OOD &\textbf{63.83} &\textbf{62.47} &\textbf{61.96} &\textbf{0.79} &\textbf{62.36} \\
\bottomrule 
\end{tabular}
\end{table*}

We report the training time of our method based on SimCLR~\citep{chen2020simple} for an epoch in~\tref{tab:training_time}, measured on a computer with two Nvidia GeForce RTX 2080 Ti GPUs and an Intel Core i9-10940X CPU. We report the processing time for updating an OOD sample set $\calD_s^{OOD}$ separately in~\tref{tab:sampling_time} because sampling is only processed every 50 epochs for the ImageNet-100-LT dataset and every 25 epochs for all other datasets. This includes ID sample clustering, tailness estimation, budget allocation, and sampling. Because we use a fixed total sampling budget ($N_b=10,000$) for all datasets regardless of the size of OOD datasets, a larger OOD dataset only increases the time for OOD data sampling while not affecting the time for other processing. Based on these reported times, the total training time for the CIFAR-10-LT dataset is calculated as follows: 9.65 (seconds/update) $\times$ 2000 (epochs) / 25 (epochs/update) + 4.32 (seconds/epoch) $\times$ 2000 (epochs) + 2.32 (seconds/epoch) $\times$ 500 (epochs), which is approximately 2.94 hours.

\begin{table}[!h]
\begin{minipage}{1\linewidth}
\centering
\caption{Analysis of the training time for an epoch in seconds.}
\label{tab:training_time}
\centering
\begin{tabular}{>{\centering}m{0.3\textwidth}| >{\centering}m{0.2\textwidth}| >{\centering\arraybackslash}m{0.2\textwidth}} 
\toprule
ID Dataset & Pre-train (Sec. 3.1) & Distillation (Sec. 3.2)     \\
\midrule
CIFAR-10-LT & 4.32 & 2.32    \\
CIFAR-100-LT & 3.87 & 1.98  \\
ImageNet-100-LT & 11.25 & 6.82  \\
Places-LT & 37.84 & 33.12   \\
\bottomrule 
\end{tabular}
\end{minipage}
\end{table}

\begin{table}[!h]
\begin{minipage}{1\linewidth}
\centering
\caption{Analysis of the time for updating $\calD_s^{OOD}$ in seconds. The CIFAR-10-LT dataset is utilized as the ID dataset.}
\label{tab:sampling_time}
\centering
\begin{tabular}{>{\centering}m{0.45\textwidth}| >{\centering\arraybackslash}m{0.3\textwidth}} 
\toprule
 OOD Dataset (Size) & OOD Sampling (Sec. 3.1)  \\
\midrule
 300K Random (300K) &  9.65  \\
 ImageNet-R  (12K) & 1.07  \\
 Places-Extra69 (62.5K) & 3.24  \\
\bottomrule 
\end{tabular}
\end{minipage}
\end{table}

\begin{table*}[!h]
\centering
\caption{Measured time for updating $\calD_s^{OOD}$ in seconds. The CIFAR-10-LT~\citep{cui2019class} dataset is utilized as the ID dataset, and the 300K Random~\citep{hendrycks2018deep} dataset is used as the OOD dataset.}
\label{tab:detailed_sampling_time}
\begin{minipage}{1\linewidth}
\centering
\begin{tabular}{>{\centering}m{0.33\textwidth}| >{\centering}m{0.1\textwidth}| >{\centering}m{0.15\textwidth}| >{\centering}m{0.1\textwidth}| >{\centering\arraybackslash}m{0.1\textwidth}} 
\toprule
Processing for updating $\calD_s^{OOD}$ (Sec. 3.1) & Clustering & Tailness estimation & Sampling & Total  \\
\midrule
Time & 0.85 & 0.06 & 8.74 & 9.65 \\
\bottomrule 
\end{tabular}
\end{minipage}
\end{table*}

Additionally, we show the detailed processing time for updating $\calD_s^{OOD}$ in~\tref{tab:detailed_sampling_time}. The time is measured using the CIFAR-10-LT dataset~\citep{cui2019class} as the ID dataset and the 300K Random dataset~\citep{hendrycks2018deep} as the OOD dataset. Note that the rightmost column of~\tref{tab:detailed_sampling_time} is the same as that of~\tref{tab:sampling_time}.

\section{CONCLUSION}
Towards robust self-supervised learning on long-tailed datasets, we present a novel method that pre-trains a network on ID and OOD data and further optimizes the network for an ID dataset. We propose to leverage the pre-trained network to discover positive/negative samples and control the strengths of attractive/repulsive forces in contrastive learning. Additionally, to maintain the balanced and well-separated embedding space from the pre-trained network, we propose to distil the embedding space and transfer to the training network. Moreover, during pre-training on ID and OOD data, we propose to assign pseudo semantic labels and encode semantic relationships to learn a balanced and well-separated embedding space. Lastly, we evaluate the proposed method using linear probing and few-shot learning protocols on four public datasets. The results demonstrate that our method outperforms previous state-of-the-art methods.

\section*{Acknowledgments}
This work was supported by the Korea Evaluation Institute of Industrial Technology (KEIT) Grant through the Korea Government (MOTIE) under Grant 20018635.

\bibliographystyle{elsarticle-harv}
\bibliography{mybibfile}

\end{document}